%% file: main.tex
\documentclass[10pt,twocolumn,letterpaper]{article}

\usepackage{cvpr}              

\input{preamble}

\definecolor{cvprblue}{rgb}{0.21,0.49,0.74}
\usepackage[pagebackref,breaklinks,colorlinks,citecolor=cvprblue]{hyperref}

\def\paperID{36} 
\def\confName{CVPR}
\def\confYear{2024}

\title{TrackMe: \\ A Simple and Effective Multiple Object Tracking Annotation Tool}

\author{Thinh Phan, Isaac Phillips, Andrew Lockett, Michael T.Kidd and Ngan Le \\
University of Arkansas \\
Fayetteville, Arkansas, USA \\
{\tt\small \{thinhp, imp002, alockett, mkidd, thile\}@uark.edu}
}

\begin{document}
\maketitle
\input{sec/0_abstract}    
\input{sec/Introduction}
\input{sec/TrackingTool}

\input{sec/Conclusion}
{
    \small
    \bibliographystyle{ieeenat_fullname}
    \bibliography{main}
}


\end{document}

%% file: preamble.tex
\usepackage[dvipsnames]{xcolor}


%% file: sec/0_abstract.tex
\begin{abstract}
Object tracking, especially animal tracking, is one of the key topics that attract a lot of attention due to its benefits of animal behavior understanding and monitoring. Recent state-of-the-art tracking methods are founded on deep learning architectures for object detection, appearance feature extraction and track association. Despite the good tracking performance, these methods are trained and evaluated on common objects such as human and cars. To perform on the animal, there is a need to create large datasets of different types in multiple conditions. The dataset construction comprises of data collection and data annotation. In this work, we put more focus on the latter task. Particularly, we renovate the well-known tool, LabelMe, so as to assist common user with or without in-depth knowledge about computer science to annotate the data with less effort. The new tool named as TrackMe inherits the simplicity, high compatibility with varied systems, minimal hardware requirement and convenient feature utilization from the predecessor. TrackMe is an upgraded version with essential features for multiple object tracking annotation. \url{https://github.com/UARK-AICV/TrackGUI}

\end{abstract}

%% file: sec/Introduction.tex
\section{Introduction}
Animal tracking is a crucial task for the purpose of studying animal behaviors and movements \cite{hertel2020guide}, bringing many benefits for promoting the natural development, and monitoring animal well-being and productivity. Tracking task on animal could be carried out in multiple scenarios (e.g. outdoor \cite{bailey2018use}, indoor \cite{colles2016monitoring}) and with various types of devices like RFID \cite{floyd2015rfid}, GPS \cite{clark2006advanced}, UWB \cite{van2019validation} and camera \cite{francisco2020high}. Although attachable monitoring devices come ready with object locations and movements, they could not provide activity or condition-related information. In contrast, camera system can meet the above requirements but depends totally on other techniques to infer the object coordinate. Nevertheless, thanks to advance in deep learning and multiple object tracking (MOT) methods, this weakness has become less significant. Recent approaches in MOT \cite{nguyen2024multi, cao2023observation, nguyen2022multi, zhang2022bytetrack, bewley2016simple, zeng2022motr} and GMOT \cite{anh2024tp, tran2023z, bai2021gmot, huang2019got} focused largely on developing models and have  achieved great performance in both object detection and object tracking. Despite the success, those methods were established on limited domains of datasets with common subject like human and car, making it hard to extend the task to animal field. As part of dataset foundation, data collection is not a big problem anymore since high-quality cameras and storage devices are widespread and affordable. On the other hand, data annotation yet demands masses of human work. 

Several track-support video annotation tools have been introduced with fascinating features. DarkLabel \cite{DarkLabel} allows bounding box annotation with object label and ID , varied data formats saving and comes with linear interpolation for multi-target box generation. DarkLabel functions are straightforward and easy to use. However, it can only run on Windows and the source code is not available to develop. CVAT \cite{boris_sekachev_2020_4009388} and LabelStudio \cite{LabelStudio} are the two of most well-built and used annotation tool for image and video. While LabelStudio features simple box linear interpolation, CVAT uses object detection model to predict the next frame's box and then matches the ID. The weakness of CVAT is that it could only detects and tracks common subjects in COCO dataset \cite{lin2014microsoft}. Irregular objects or untrained types of animal would lead to bad result. In addition, the two annotation tools are built for wide range of image and video labeling tasks, making it complicated for new users to set up the new project and get accustomed to the labeling procedure. 

\begin{figure*}
    \centering
    \includegraphics[scale=0.55]{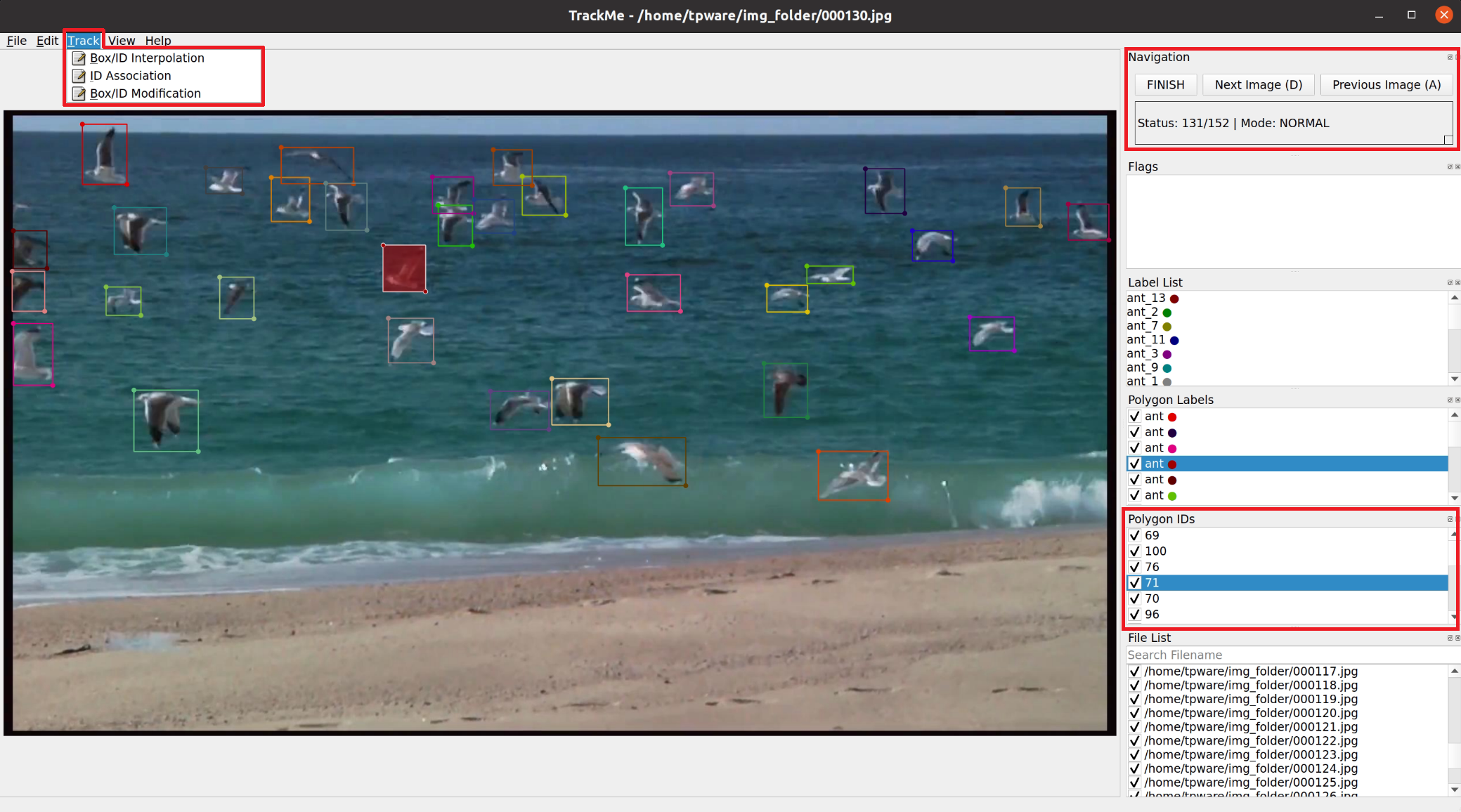}
    \vspace{-1em}
    \caption{The overview of the graphical interface of TrackMe. The red boxes highlights the new features and utilities. We add "Track" in the menu list along with three video tracking annotation features. The "Navigation" dialog is placed on the top right for frame control. "Polygon IDs" demonstrates the IDs of current frame's objects. The images are from GMOT-40 \cite{bai2021gmot}.}
    \label{fig:TrackMe_overview}
\end{figure*}

In this paper, we introduce a new video annotation tool, called TrackMe, mainly built for MOT tracking. Our tool is based on LabelMe \cite{wada2018labelme}, which is a popular tool for multi-purposed image annotation. LabelMe is selected because of several beneficial factors of tool development. It is an open source tool and the code is relatively straightforward for developers. The code is Python-based so that the software environment installation would be uncomplicated and the tool is compatible on any OS. The hardware requirement for LabelMe is trivial along with the small computation. In terms of user experience, LabelMe added sufficient features for image annotation and its graphical interface are quite intuitive to use. TrackMe inherits all the advantages from LabelMe and is integrated with fundamental tracking annotation features such as bounding box interpolation, ID association, and info modification. LabelMe lacks the ability to generate and revise the bounding boxes in continuous frames. While bounding box interpolation reduces the manual labeling of individual object's boxes in every frames, association feature keeps track of multiple objects and automatically assigns ID for them. Should the boxes of one object are wrong, TrackMe is able to adjust a series of images with one function. 

Our goal is to provide a new open source tool that could facilitate the dataset annotation for users of wide range of computer knowledge levels. The detailed tool description in this paper could, to some extent, guide other developers on how to further improve the tool with new tracking features.

%% file: sec/TrackingTool.tex
\section{Tracking Tool}
\label{sec:trackingtool}

\subsection{LabelMe Annotation Tool}
\label{sec:sub1}
LabelMe \cite{wada2018labelme} is a well-known annotation tool owing to the diverse annotation styles (e.g. polygon, rectangle, keypoints), efficiency and user-friendly interface and features. For instance, in terms of object detection, users choose "Create Rectangle" in the "Edit" menu, draw the bounding box and input the object category. The frame annotation file is saved as dictionary with fundamental keys such as image name, object info, annotation types and points. LabelMe currently does not have annotation key features and cannot save or edit object ID key. The bounding boxes are depicted with identical colors  with objects of same classes, which is not ideal for visually following adjacent and moving objects. "Keep Previous Annotation" could slightly be acceptable as the tracking feature for stationary objects because it simply copies the boxes from current frame to next frame. In the coming sections, we will discuss how to capitalize the aforementioned characteristics for TrackMe.


\subsection{Tracklet Information and Visualization}
\label{sec:sub2}
\cref{fig:TrackMe_overview} depicts the TrackMe tool and the red boxes highlights the newly-developed functions. LabelMe contains a "Polygon Labels" dialog that shows all object label names and a "Label List" dialog that retain the non-overlapping ones. The unique box colors are assigned based on the non-overlapping labels. Based on this, firstly, we include the object ID field in the object info and we also create the "Polygon IDs" dialog with same coding structure as "Polygon Labels" to list all object IDs in the frame. Instead of only feeding the labels, the string combination of object category and ID is provided to "Label List", making individual objects be viewed with different colors and easier to follow. TrackMe highlights the clicked object and its label and ID info, too. A "Navigation" dialog is placed on the top right so that users are aware of the position of the current frame and the total number of frames in the video folder, the editing status and a confirmation button for further features. "Track" is introduced in the menu with three main video annotation functions, which are discussed in the below sections \ref{sec:sub3}, \ref{sec:sub4}, \ref{sec:sub5}.

\subsection{Box \& ID Interpolation}
\label{sec:sub3}

\begin{figure}[!h]	
    \centering
    \includegraphics[width=0.5\linewidth]{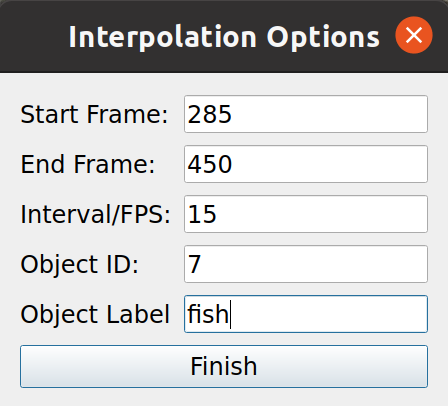}
    \caption{Box and ID Interpolation Dialog.}
    \label{fig:box_interpolation}
\end{figure}

To help users cut down the workload when annotating same object's boxes over lengthy continuous frames, we integrated "Box/ID Interpolation" feature into LabelMe. The interpolation graphical interface is shown in \cref{fig:box_interpolation}. Given a sequence of frames missing object boxes, the idea is that users demonstrate a few boxes in some certain frames and interpolation feature will complete the rest. We construct this feature based on Gaussian Process Regression (GPR) \cite{seeger2004gaussian} technique due to its effectiveness and flexibility in continuous data regression issues. Since object movement is non-linear through time, the non-parametric nature of GPR plays a key role in adjusting the function itself to the data complexity. Additionally, the generated data is less prone to noisy input and pleasantly fits the real pattern. We employ GPR with Rational Quadratic Kernel. The function input is the normalized frame orders (temporal data) and the output is the box coordinate under the format of (center x, center y, width and height). GPR is trained with data from selectively labeled frames and then fills out missing boxes on the others. The training and prediction are quick and requires no GPU.

In practice, user inputs the following information: start and end frame numbers, interval length, object label and ID. The interpolation is only executed between start and end frames and produces boxes with input label and ID. If ID is left blank, the generated boxes comes with ID as null. Interval value decides the evenly spaced frames for box guidance. Video frame per second normally works good as interval value. Interval number should be set smaller to compensate the movement error in case object has plenty of movements in that duration or larger if the object movements are inconsiderable. In normal mode, users can search and edit any files in the "File List" dialog. In interpolation mode, TrackMe prevents unnecessary labeling by merely let users reach the chosen ones. After boxes are drawn on those frames, users press the "Finish" button in the "Navigation" dialog. Eventually, the boxes with specified label and ID are updated in the remaining frames.

\subsection{ID Association}
\label{sec:sub4}

\begin{figure}[!h]	
    \centering
    \includegraphics[width=0.5\linewidth]{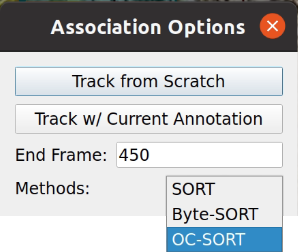}
    \vspace{-1em}
    \caption{ID Association Dialog.}
    \label{fig:association_dialog}
\end{figure}

Supposed that objects in the video are already covered with bounding boxes except IDs, "Track Association" is the ideal feature for automatic ID assignment. This feature is inspired by tracking-by-detection methods where the objects are initially localized and bounding boxes in adjacent frames are matched by comparing the spatial proximity, motion inclination and appearance features. Some of the well-known methods are SORT \cite{bewley2016simple}, ByteTrack \cite{zhang2022bytetrack}, OC-SORT \cite{cao2023observation} and MOTR \cite{zeng2022motr}. LabelMe saves frame labels inside .json (dictionary) file with well-defined structure, which is handy to both derive and modify the information. Pretrained deep learning object-related detection model are utilized to predict the boxes and newly create label files for all the frames. Subsequently, association method is applied to fill out the IDs. In TrackMe, we include the bounding box training and prediction code based on YOLO-v8 \cite{Jocher_Ultralytics_YOLO_2023} and box association based on SORT. Since we follow the input format shared by most of association methods, it is convenient to incorporate state-of-the-art ones into TrackMe later on.

\begin{figure*}[!t]
    \centering
    \includegraphics[width=\linewidth]{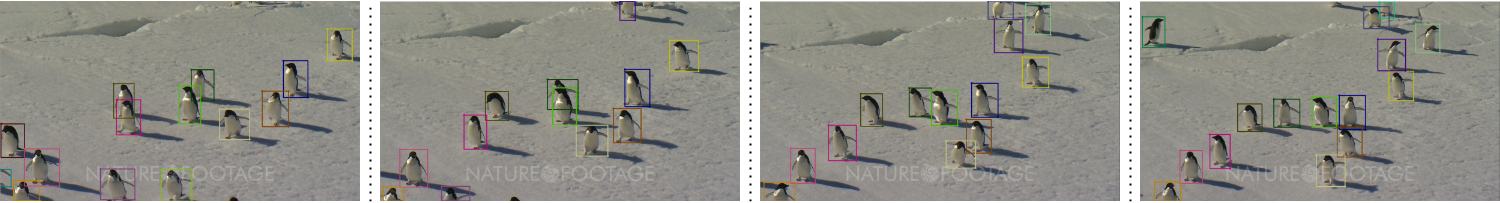}
    \vspace{-2em}
    \caption{Tracklet Annotation Visualization in TrackMe. The images are from GMOT-40 \cite{bai2021gmot}.}
    \label{fig:tracking sample}
\end{figure*}

TrackMe is built with two association features as displayed in \cref{fig:association_dialog}. "Track from Scratch" randomly assigns the IDs for the objects in the first frame and propagates them to the following frames. In case users need to manually place the IDs onto the objects, "Track with Current Annotation" uses the already-assigned IDs in the current frame and spreads it forward. This feature is also a way to correct the wrongly-assigned IDs. Users scroll to the erroneous frame, rectify the IDs and choose the second option. The previous frames' annotation are kept intact and the following frames are modified according to current frame's information. "End frame" can be left blank if users wants to complete the whole video or set with a number to interrupt the program. There are several association methods users can select.

\subsection{Box \& ID Modification}
\label{sec:sub5}

\begin{figure}[!h]	
    \centering
    \includegraphics[width=0.5\linewidth]{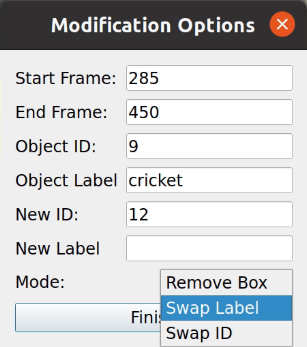}
    \vspace{-1em}
    \caption{Box and ID Modification Dialog.}
    \label{fig:modification_dialog}
\end{figure}

Box interpolation technique and ID assignment model may sometimes be flawed throughout a sequence of images but LabelMe is restricted to singular frame correction. Hence, "Box \& ID Modification" is featured in TrackMe with the capability of box deletion, and label and ID adjustment. The feature dialog as shown in \cref{fig:modification_dialog} has some similar fields as the "Box \& ID Interpolation" dialog, replaces interval number with new ID value and new label value, and adds modification mode. Likewise, users fills out the frame range, box ID and label. For removal features, new ID or label value does not need to be filled. Mode "remove all" activates both box and ID deletion. Mode "swap ID" and "swap label" will replaces the corresponding old info of the object with new one. TrackMe will loop through the picked frames and only modify the pertained boxes.

%% file: sec/Conclusion.tex
\section{Conclusion}
In this paper, we present TrackMe, an easily operated and comprehensive video tracking tool for any subjects. The tool is constructed with the aim to help more people, with or without computer science knowledge, comfortably install and utilize the tracking features. TrackMe adds ID field to LabelMe, adapts the graphical interface for object tracking, and integrated track generation and modification features. TrackMe will be further updated with new video labeling features and the code will be published for other developers to polish the tool. 